\PassOptionsToPackage{table}{xcolor}
\documentclass[11pt,a4paper]{article}
\usepackage[final]{style/neurips_2021}
\usepackage{times}
\usepackage{latexsym}
\usepackage{enumitem}

\newif\ifcomment\commentfalse

\usepackage[a-1b]{pdfx}

\usepackage{framed}
\usepackage{mdwlist}
\usepackage{siunitx}
\usepackage{latexsym}
\usepackage{colortbl}
\usepackage{xcolor}
\usepackage{nicefrac}
\usepackage{booktabs}
\usepackage{fnpct}
\usepackage{amsfonts}
\usepackage[T1]{fontenc}
\usepackage{bold-extra}
\usepackage{amsmath}
\usepackage{amssymb}
\usepackage{bm}
\usepackage{graphicx}
\usepackage{mathtools}
\usepackage{microtype}
\usepackage{multirow}
\usepackage{multicol}
\usepackage{xpatch}
\usepackage{latexsym,comment}
\usepackage[normalem]{ulem}

\newcommand*{\missingreference}{{\Huge \colorbox{red}{?reference?}}}
\newcommand*{\missingcitation}{{\Huge \colorbox{red}{?citation?}}}

\makeatletter
\xpatchcmd{\@setref}{\bfseries}{\missingreference}{}{}
\def\@citex[#1]#2{\leavevmode
    \let\@citea\@empty
    \@cite{\@for\@citeb:=#2\do
        {\@citea\def\@citea{,\penalty\@m\ }%
            \edef\@citeb{\expandafter\@firstofone\@citeb\@empty}%
            \if@filesw\immediate\write\@auxout{\string\citation{\@citeb}}\fi
            \@ifundefined{b@\@citeb}{\hbox{\reset@font\missingcitation}%
                \G@refundefinedtrue
                \@latex@warning
                {Citation `\@citeb' on page \thepage \space undefined}}%
            {\@cite@ofmt{\csname b@\@citeb\endcsname}}}}{#1}}
\makeatother

\newcommand{\gem}[1]{\mbox{\textsc{gem}}}
\newcommand{\abr}[1]{\textsc{#1}}
\newcommand{\camelabr}[2]{{\small #1}{\textsc{#2}}}

\newcommand{\hidetext}[1]{}
\newcommand{\ignore}[1]{}

\ifcomment
    \newcommand{\pinaforecomment}[3]{\colorbox{#1}{\parbox{.8\linewidth}{#2: #3}}}

    \newcommand{\prtodo}[1]{\pinaforecomment{lightblue}{pr}{#1}}
    \newcommand{\prtodoi}[1]{\pinaforecomment{lightblue}{pr}{#1}}
\else
    \newcommand{\pinaforecomment}[3]{}
    \newcommand{\prtodo}[1]{}
    \newcommand{\prtodoi}[1]{}
\fi

\newcommand{\smallurl}[1]{ \begin{tiny}\url{#1}\end{tiny}}

\definecolor{lightblue}{HTML}{3cc7ea}
\definecolor{CUgold}{HTML}{CFB87C}
\definecolor{grey}{rgb}{0.95,0.95,0.95}
\definecolor{ceil}{rgb}{0.57, 0.63, 0.81}
\definecolor{UMDred}{HTML}{ed1c24}
\definecolor{UMDyellow}{HTML}{ffc20e}

\usepackage{booktabs}
\usepackage{cleveref}
\usepackage{soul}
\usepackage{caption}
\usepackage{subcaption}

\title{Is Automated Topic Model Evaluation Broken?: \\ The Incoherence of Coherence }

\author{
    \normalfont
    \noindent\makebox[0pt]{
    \begin{tabular}{@{}c c c c@{}}
        \textbf{Alexander Hoyle}\thanks{Equal contribution} & \textbf{Pranav Goel}\footnotemark[1] & \textbf{Denis Peskov}\footnotemark[1] & \textbf{Andrew Hian-Cheong}\footnotemark[1]\\
        \multicolumn{4}{c}{Computer Science}\vspace*{0.5em} \\
        & \textbf{Jordan Boyd-Graber} & \textbf{Philip Resnik} & \\
        & \abr{cs}, iSchool, \abr{umiacs}, \abr{lsc} & \abr{umiacs}, Lingusitics & \vspace*{0.5em} \\
        \multicolumn{4}{c}{University of Maryland} \\
        \multicolumn{4}{c}{\texttt{\{hoyle,pgoel1,dpeskov,andrewhc,jbg,resnik\}@cs.umd.edu}}
    \end{tabular}
    }
}

\newcommand{\figfile}[1]{2021_neurips_topics/figures/#1}

\newcommand{\intrusion}{word intrusion}
\newcommand{\ratings}{ratings}

\begin{document}
\maketitle

\begin{abstract}
  Topic model evaluation, like evaluation of other unsupervised methods, can be contentious.
    However, the field has coalesced around automated estimates of topic coherence, which rely on the frequency of word co-occurrences in a reference corpus.
     Contemporary neural topic models surpass classical ones according to these metrics. 
      At the same time, topic model evaluation suffers from a \emph{validation gap}: automated coherence, developed for classical models, has not been validated using human experimentation for neural models.
        In addition, a meta-analysis of topic modeling literature reveals a substantial \emph{standardization gap} in automated topic modeling benchmarks.
        To address the validation gap, we compare automated coherence with the two most widely accepted human judgment tasks: topic rating and word intrusion.  
       To address the standardization gap, we systematically evaluate a dominant classical model and two state-of-the-art neural models on two commonly used datasets.
      Automated evaluations declare a winning model when corresponding human evaluations do not, calling into question the validity of fully automatic evaluations independent of human judgments.\looseness=-1
  \end{abstract}

\section{Revisiting Topic Model Evaluation}
\label{sec:intro}

Topic models are a machine learning technique widely used outside computer science, including political science~\citep{political,Karoliina2021Topic}, social and cultural studies~\citep{MOHR2013545}, digital humanities~\citep{meeks2012digital}, and bioinformatics~\citep{Liu2016AnOO}. 
Typically, topic model users are domain experts trying to identify global categories or themes present in a document collection~\citep{boyd-graber-17}. 
This practice constitutes a computer-assisted form of content analysis~\citep{kripp2004,Chuang2014ComputerAssistedCA}, also related to distant reading in literary studies~\citep{underwoodGenealogy2017}.
In general, topic models help humans understand large corpora.\footnote{Topic models are also used for other purposes, such as information retrieval or downstream document classification. 
However, the discovery and application of categories for human interpretation is their dominant use, and other computational applications have been largely eclipsed by modern neural approaches.}\looseness=-1

\begin{table}[t]
	\small
	\begin{center}
		\begin{tabular}{ l l l l | l l l }
			\toprule
			& \multicolumn{3}{c}{\textbf{Classical}}& \multicolumn{3}{c}{\textbf{Neural}} \\
			\midrule
			  & station               & album                 & tropical              & tropical              & spore                 & manhattan\_project    \\ 
			  & line                  & band                  & storm                 & landfall              & basidia               & los\_alamos\_laboratory\\ 
			  & bridge                & music                 & hurricane             & cyclone               & spores                & robert\_oppenheimer   \\ 
			  & railway               & song                  & cyclone               & utc                   & mycologist            & enrico\_fermi         \\ 
			  & trains                & released              & depression            & weakening             & hyphae                & physicist            \\
				\midrule
			\textbf{NPMI} & 0.274 & 0.285 & 0.394 & 0.446 & 0.456 & 0.470 \\
			\bottomrule
		\end{tabular}
	\end{center}
	\caption{
				The first three columns are the highest-\abr{npmi} topics for a classical topic model~\citep[\abr{lda}
		estimated via Gibbs sampling using
		Mallet,][]{McCallumMALLET, Griffiths2004FindingST}. The next three are counterparts from a
		neural model~\citep[our \abr{d-vae}
		reimplementation,][]{burkhardtDecouplingSparsitySmoothness2019}.
		Models are trained on
		Wikitext~\citep{merityPointerSentinelMixture2017} with fifty
		topics, and \abr{npmi} is estimated over the top five words
		in each topic using a 4.6M-document reference Wikipedia
		corpus. The mean top-five \abr{npmi} over all topics is 0.156
		for the classical and 0.256 for the neural model. 	}
	\label{tab:topic_ex}
\end{table}

Evaluation of topic models has vacillated between automated and human-centered.
While real-world users of topic models evaluate outputs based on their specific needs, topic model developers have gravitated toward generalized, automated proxies of human judgment to help inform rapid iteration of models~\citep[][]{doogan-buntine-2021-topic}.
Initially, models were evaluated with held-out perplexity, but it disagrees with human interpretability \citep[][]{chang2009reading}.
Consequently, the field adopted automated coherence metrics like normalized pointwise mutual information (\abr{npmi}), a measure of word relatedness that \emph{does} correlate with topic interpretability~\citep[Section~\ref{sec:bg:evaluation};][]{newman2010automatic,aletras2013evaluating,lau2014machine}.
The balance shifted towards automated coherence.

Human evaluations have been abandoned by topic model developers in the years since automated coherence metrics were adopted.
In a thorough meta-analysis of contemporary topic model methods papers, \emph{none} conduct systematic human evaluations (Section~\ref{sec:meta}).
Instead, they rely solely on automated metrics for model comparison.\footnote{Outside of the core method-development literature, human evaluations have been used to develop new metrics and improve understanding of existing model behavior \citep[][\emph{inter alia}]{bhatia-etal-2017-automated,morstatter2018search,lund-etal-2019-automated,alokaili-etal-2019-ranking}.}
However, current neural topic models are a far cry from the classical models that substantiated the original correlations---manifestly, topics produced by neural models are often qualitatively distinct from those of classical models (e.g., Table~\ref{tab:topic_ex}).\footnote{We use ``classical'' to mean generative models defined by a chain of conjugate exponential family distributions optimized by Gibbs sampling or variational inference.} 
This \emph{validation gap} raises the question of whether automated metrics are still consistent with human judgments of topic quality.

Moreover, we should always be cautious when extrapolating outside the range of data that was used to establish a relationship between variables.
As an example, a neural model in \citet{hoyle-etal-2020-improving} produces much larger \abr{npmi} values than those used to determine human correlations in the original~\citet{lau2014machine} study; the implicit assumption is that greater \abr{npmi} corresponds to more human-interpretable topics.
Finally, a myopic focus on a presumed proxy for human preferences can produce low-quality results~\citep{Stiennon2020LearningTS}. 
Does Goodharts' law---``when a measure becomes a target, it ceases to be a good measure''~\citep{Strathern1997ImprovingRA}---apply to automated metrics of topic models?\looseness=-1

Another challenge for automated evaluation, whether of classical or neural topic models, is widespread inconsistency (Section~\ref{sec:meta}).
Researchers frequently fail to specify the information needed to
calculate automated metrics or diverge from the practices
that underpin human correlations.
Furthermore, evaluation datasets, preprocessing, and hyperparameter
optimization vary dramatically, even within a given paper.
This \emph{standardization gap} likely limits the generalizability and reliability of topic model developers' findings.

We address the standardization and validation gaps in topic model evaluation:\begin{enumerate*}
	\item{We present a meta-analysis of neural topic model evaluation (Section~\ref{sec:meta})};
	\item{we develop standardized, pre-processed versions of two widely-used English-language evaluation datasets, along with a transparent end-to-end code pipeline for reproduction of results (Section~\ref{sec:models:preprocessing})\footnote{\url{github.com/ahoho/topics}}};
	\item{we optimize three topic models---one classical and two neural---using identical preprocessing, model selection criteria, and hyperparameter tuning (Section~\ref{sec:models:models});}
	\item{we evaluate these models using human \ratings{} and \intrusion{} tasks (Section~\ref{sec:human_exp_setup}); and}
	\item{we provide new evaluations of the correlation between automated and human evaluations (Section~\ref{sec:analysis}).}
\end{enumerate*}

Our findings challenge the validity of fully-automated evaluations as currently practiced: automated evaluation declares winners between models when the corresponding human evaluations cannot.
\section{Operationalizing Topic Coherence}
\label{sec:bg}

A topic model is a probabilistic generative model of text that uses
latent \emph{topics} to summarize a larger collection of documents.
The most influential variant, latent Dirichlet
allocation~\citep[\abr{lda}]{bleiLatentDirichletAllocation2003}, assumes that
$K$~latent topics are distributions over word types, $\bm{\beta}_k$, and
that the documents $\mathcal{D}$ are admixtures over the topics,
$\bm{\theta}_d$.
Users often evaluate model outputs globally, focusing on the most
probable~$N$ words of each topic, and locally, considering
the most probable topics for each document.

While techniques for topic modeling have progressed from variational
inference~\citep{bleiLatentDirichletAllocation2003} to Gibbs
sampling~\citep{Griffiths2004FindingST} to deep generative
approaches~\citep{Srivastava2017AutoencodingVI,Wang2019ATMAT}, the
core goal discussed in Section~\ref{sec:intro}, obtaining human-understandable categories, remains central.
The latest wave of methods, \emph{neural topic models}
(\abr{ntm}), use continuous word representations and
gradient optimization to fit parameters. 
These models claim to produce more interpretable topics than other prior methods, including  \abr{lda}.

Those claims are supported by improvements on automated measures of topic coherence.

\subsection{Human Metrics of Topic Coherence}\label{sec:bg:human-eval}

Like the concept of \emph{interpretability}, that of real-world \emph{coherence} is ``simultaneously important and slippery''~\citep{lipton2018mythos}.  We will not attempt to formalize it here---though see discussion in Section~\ref{sec:conclusion}.
For present purposes, the term has its roots in Latin \emph{cohaerere}, ``to stick together,'' and we will think of coherence as an intangible sense, available to human readers, that a set of terms, when viewed together, enable human recognition of an identifiable category.\footnote{This perspective aligns with Propositions 2 and 3 of~\cite{doogan-buntine-2021-topic}: ``an interpretable topic is one that can be easily labeled,'' and ``has high agreement on labels.''}
We review two human ratings of topic quality: direct ratings and intrusion.

\paragraph{Rating}
Raters see a topic and then give the topic a
quality score, conventionally on a three-point ordinal scale~\cite[\emph{inter
  alia}]{newman2010automatic, mimno2011optimizing,
  aletras2013evaluating}.

\paragraph{Intrusion}
\citet{chang2009reading} devise the \emph{word intrusion} task as a behavioral way to assess topic coherence. The core idea is that when the top words in a topic identify a coherent latent category, it is easier to identify words that do not belong to that category.
Operationally, each topic is represented as its top words plus one ``intruder'' word
which has a low probability of belonging to that topic, but a high
probability of belonging to a different topic.
Topic coherence is then judged by how well human annotators detect the
``intruder'' word.

\subsection{\abr{npmi}: The Standard Automated Topic Model Coherence Evaluation}
\label{sec:bg:evaluation}

Using the \intrusion{} task, \citet{chang2009reading} showed that perplexity---the original topic model evaluation metric---\emph{negatively} correlates with human evaluations of topic quality.
This finding revealed a need for an automated measurement of topic coherence:  
an automated metric can measure model quality 
without expensive, time-consuming, and difficult-to-reproduce human experiments.

\citet{lau2014machine} find some metrics that \emph{positively} correlate with human intrusion and rating scores, particularly when aggregating scores over all topics from a given model.
Because of that validation, the prevailing evaluation for model comparison is pairwise normalized pointwise mutual information.
\abr{npmi} scores topics highly if the
top~$N$ words---summed over all pairs $w_i$ and $w_j$---have high
joint probability $P(w_j, w_i)$ compared to their marginal
probability:\footnote{Alternative metrics
exist, but they typically also rely on either joint probability
estimates or \abr{npmi} directly~\citep[e.g.,
  $C_v$][]{roder2015exploring}.}
\begin{equation}
   \sum_{j=2}^{N} \sum_{i=1}^{j-1} \frac{log \frac{P(w_j, w_i)}{P(w_i) P(w_j)}}{- log P(w_i, w_j)}.
\label{eq:npmi}
\end{equation}
The probabilities are estimated using word co-occurrence counts from a
\emph{reference corpus} for a specific context window (which can range
from ten words to the entire document). 
As a result, the choice of reference corpus determines the strength of
human correlation~\citep{lau2014machine,roder2015exploring}.

A measurement is \emph{valid} to the extent that it measures what it is intended to measure in the real world.
Historically, automated coherence has been validated using \emph{human} judgements from either crowdworkers~\citep{newman2010automatic,aletras2013evaluating} or experts~\citep{mimno2011optimizing}.
However, correlations based on classical models may not be applicable for \abr{ntm}s.
Our skepticism is motivated by theory, as neural word
representations are intimately connected to \abr{npmi}, as explicitly
used by~\citet{aletras2013evaluating} and which produce similar \abr{npmi} scores as~\citet{lau2014machine}.
\citet{NIPS2014_feab05aa} show that multiple representations create
factorizations of \abr{pmi} matrices.
Topic models that have access to these rich representations
\citep[e.g.][ and others]{diengTopicModelingEmbedding2019} could thus create
topics with good \abr{npmi} scores without explaining the corpus well
to a user.
In contrast to classical topic models, no one has investigated the
validity of \abr{npmi} evaluation for \abr{ntm}s.\looseness=-1

Given this lacuna, we conduct experiments aimed at validating that automated topic evaluations still
correlate with human judgments of neural topic model quality.
We compare against two common human evaluations of individual topic quality: direct rating and intrusion.
Human evaluations, like automated topic modeling, lack standardization, which we address in Section~\ref{sec:human_exp_setup}.

\section{A Meta-Analysis of Neural Topic Modeling}
\label{sec:meta}
\begin{table}[t!]
  \footnotesize
  \begin{subtable}{\textwidth}
  \begin{minipage}[t]{.5\textwidth}
      \vspace{0pt}
      \begin{center}
    \begin{tabular}{ l r r }
      \toprule
    \textbf{Evaluation} &  \multicolumn{2}{l }{\textbf{Count}} \\
      \midrule
      Number of human evaluations & 0 & (0\%) \\
    \emph{Automated Coherence} \\
    \quad \emph{Metric} \\
    \quad \quad \abr{npmi} & 26 & (72\%) \\
    \quad \quad Other & 22 & (61\%) \\
    \quad Explicit implementation & 22 & (61\%) \\
    \quad Explicit ref. corpus & 10 & (28\%) \\
    \quad Perplexity w/o coherence & 3 & (8\%) \\
    \phantom{line}\\
    \bottomrule
  \end{tabular}
\end{center}
\label{tab:meta_analysis}
  \end{minipage}
  \begin{minipage}[t]{.5\textwidth}
  	\vspace{0pt}
  	\begin{center}
      \begin{tabular}{ l r r }
        \toprule
  
        \textbf{Experimentation} &  \multicolumn{2}{l }{\textbf{Count}} \\
      \midrule
        \emph{Preprocessing} \\
        \quad Inconsistent over datasets & 12 & (30\%) \\
        \quad Ambiguous preprocessing & 9 & (23\%) \\
        \emph{Model comparisons} \\
        \quad All models tuned  & 5 & (13\%) \\
        \quad Unclear h.param search & 16 & (40\%) \\
        \quad Unclear \abr{lda} baseline, if used & 7 & (24\%) \\
        \quad Recent baseline (w/in 2 yrs) & 31 & (78\%) \\
        \quad Multiple runs / sig. testing & 11 & (28\%) \\
    \bottomrule
    \end{tabular}
  \end{center}

  \end{minipage}
 \end{subtable}
\caption{Meta-analysis of forty neural topic modeling
	papers (denominator may change, as not all conditions are applicable).  No recent neural
	topic modeling papers use human evaluations of coherence, and the
	metrics and models are difficult to replicate.}
\label{tab:meta_analysis}
\end{table}

We survey the neural topic modeling (\abr{ntm})
literature to assess the state of evaluation in contemporary topic model development.
First, we take all references made by an existing, comprehensive
survey of \abr{ntm}s~\citep{Zhao2021TopicMM}, from which we select (a)
modeling papers which (b) mention topic
interpretability and (c) compare models' topics with an existing
baseline. 
This yields forty models, which all claim superior topic coherence. 
We examine data processing steps, hyperparameter tuning,
baseline selection, and automated coherence calculations.
Table~\ref{tab:meta_analysis} summarizes our results and Appendix~\ref{app:meta} enumerates the papers.

Our analysis reveals variance in all areas.
Preprocessing, which can significantly affect model quality and automated metrics, is often (30\%) inconsistent across datasets within the same paper.
When preprocessing \textit{is} consistent, authors omit details
necessary to fully replicate the pipeline.
These issues imply that automated metrics for the same baselines and source
datasets vary across papers.
Compounding the problem, researchers often train their models on different datasets from
those used to establish the relationships between human annotations and automated metrics;
\citet{doogan-buntine-2021-topic} find that the same
metrics may not predict interpretability in new domains.
Mirroring findings from~\cite{dodge-19}, 40\% of
papers fail to clearly specify their model tuning procedure, often even the metric used for model selection.

Calculation of automated coherence metrics is equally fraught. 
As discussed in Section~\ref{sec:bg:evaluation}, a complete specification for \abr{npmi} involves several pieces of information, including the reference corpus used to estimate joint word probabilities, the co-occurrence window size, and the number of words selected from the head of the topic distribution.
Three out of four papers fail to explicitly indicate the reference corpus; even when we can assume the input corpus is used (13 cases), it remains uncertain whether authors use, e.g., a held-out set or the training documents themselves.
For the 61\% that specify the implementation of
their coherence metric (by pointing to a code repository
or writing out the formula), some of these factors may still be in
question.
For instance, six authors reference~\citet{lau2014machine} and the
supporting code,\footnote{\texttt{github.com/jhlau/topic\_interpretability}} but
the implications are ambiguous: the original paper suggests a large corpus from
the same source as the training data, but the repository script
defaults to Wikipedia.
In other cases, authors use bespoke implementations, which creates room for errors, or deviate from the settings used in human experiments.
For example, several papers use a document-wide context window with \abr{npmi}, which has not been correlated with human judgments.

Last, \emph{even if} automated evaluations are consistent, all claims of coherence improvement depend on the validity results in \citet{lau2014machine} generalizing to neural topic models.
\section{Closing the Standardization Gap for Topic Models}\label{sec:models}

Our human evaluation of topic model outputs serves multiple purposes: (a) establishing whether \abr{ntm}s show improved coherence over a classical baseline and (b) re-evaluating the efficacy and reliability of automated coherence metrics.
In addition, a key goal is (c) to provide a standardized preprocessing pipeline to support head-to-head comparisons as new methods are developed.\footnote{Our preprocessing pipeline is agnostic to dataset and easily portable. \url{github.com/ahoho/topics}}

We identify two commonly-used datasets, which we in turn process using a standard pipeline.
We then estimate topic models on each dataset following a computationally fair hyperparameter search.
Our standardization efforts are similar to concurrent work by~\cite{terragni2020octis}; the main differences are that we (a) mandate consistent preprocessing between training and reference corpora, (b) support multi-word expressions during vocabulary creation (see below), and (c) support distributed hyperparameter searches.

\subsection{Datasets and Preprocessing}\label{sec:models:preprocessing}

Following~\cite{chang2009reading}, we use English articles from Wikipedia and the New York \emph{Times} (Table~\ref{tab:data}).
For Wikipedia, we use Wikitext-103
\citep[\abr{wiki},][]{merityPointerSentinelMixture2017}, and for the
\emph{Times}, we subsample roughly 15\% of documents from LDC2008T19
\citep[\abr{nyt},][]{sandhaus2008new}, making it an order of magnitude
larger than \abr{wiki}.
To compute reference counts, we use a 4.6M document Wikipedia dump from September 2017 and the full 1.8M document LDC2008T19 set, processed identically to the training data.

We use SpaCy~\citep{spacy} to tokenize and identify entities in the text.
We create new tokens for detected entities of the form \texttt{New\_York\_City}, per~\cite{krasnashchok2018improving}.
\cite{Schofield2016ComparingAT} find that lemmatization and word-stemming can hurt English topic interpretability, so we do not lemmatize.
To maintain a roughly equal vocabulary size over datasets, we use a power-law relationship of corpus size~\citep[c.f.][]{Zipf49} to rule out tokens occurring in fewer than a given number of documents.\footnote{We target vocabularies approximating the number of words known by an adult English-speaker~\citep{brysbaert2016words}: roughly 40k for \abr{Wiki} and 35k for \abr{nyt}.}
In addition to a standard stopword list, we define corpus-specific stopwords as tokens appearing in more than 90\% of documents. See Appendix~\ref{app:preprocess} for complete preprocessing details.

\subsection{Models}\label{sec:models:models}

We evaluate one venerable classical model and two newer neural models:
\paragraph{Gibbs-\abr{lda}}
As a strong classical baseline, we use the widely-loved Mallet
\citep{McCallumMALLET} implementation of Gibbs-sampling for \abr{lda}
\citep{Griffiths2004FindingST}. Mallet produces topics of (qualitatively) competitive quality to neural models~\citep{Srivastava2017AutoencodingVI}.  

\paragraph{Dirichlet-\abr{vae}}
We reimplement Dirichlet-\abr{vae}~\citep{burkhardtDecouplingSparsitySmoothness2019}, a state-of-the-art \abr{ntm}.
For simplicitly, we use pathwise gradients for the Dirichlet~\citep{Jankowiak2018PathwiseDB}, rather than the rejection sampling variational inference of the authors' primary variant.\footnote{We replicate their \abr{npmi} and redundancy scores on 20 newsgroups. \url{github.com/ahoho/dvae}}
Dirichlet-\abr{vae} is a wholesale improvement on one of the first successful \abr{ntm}s, the popular \camelabr{Prod}{lda}~\citep{Srivastava2017AutoencodingVI}, and is competitive against recent models on automated coherence.
The generative model is simple and retains a broad similarity to \abr{lda}. The primary difference is that it does not constrain the estimated topic-word distributions to the simplex.

\paragraph{\abr{etm}} 
Thanks to their improved flexiblity, many \abr{ntm}s
incorporate external word representations, on the premise that large-scale, general language knowledge improves topic quality~\citep{Bianchi2020PretrainingIA,hoyle-etal-2020-improving}.
The Embedded Topic Model~\citep{diengTopicModelingEmbedding2019} is a
popular \abr{ntm} that relies on word embeddings in its
generative model.\footnote{\url{github.com/adjidieng/ETM}}
We maintain a fixed computational budget per model following the exhortation of~\cite{dodge-19} and use a random set of
164 hyperparameter settings across datasets for each model type.\footnote{While runtimes can vary drastically by model, this study is not concerned with implementation efficiency~\citep[although efficiency matters, see][]{ethayarajh2020utility}.}
We train models for a variable number of steps (a hyperparameter); to calculate automated coherence for the model, we use the topics produced at the last step.
For human evaluations, we select the models that maximize \abr{npmi}, estimated using the reference corpus with a ten-word window over the top ten topic words, per~\citet{lau2014machine}.
We follow the recommendation of~\cite{diengTopicModelingEmbedding2019} and learn skip-gram embeddings on the training corpus for \abr{etm} (experiments with external pretrained embeddings did not yield substantially different results).
As in \cite{hoyle-etal-2020-improving}, we eliminate models with highly redundant topics, a known degeneracy of \abr{ntm}s~\citep{burkhardtDecouplingSparsitySmoothness2019}: (a) models in which any of the top five words of one topic overlap with another and (b) models that have a topic uniqueness score~\citep{Nan2019TopicMW} above 0.7.
Ranges for hyperparameters and other details are in Appendix~\ref{app:model-detail}.

\begin{figure}[t!]
	\centering
		\includegraphics[width=0.8\textwidth]{\figfile{intrusion.pdf}}
	\caption{The \intrusion{} task presented to crowdworkers (the \ratings{} task is in Appendix~A.\ref{fig:tasks:ratings}).}
	\label{fig:tasks}\label{fig:tasks:intrusion} 
\end{figure}

\section{Human Evaluations of Topic Quality}\label{sec:human_exp_setup}

We use the \emph{ratings} and \emph{word intrusion} tasks from Section~\ref{sec:bg:evaluation} as human evaluations of topic quality.
We recruit crowdworkers using Prolific.co, an online panel provider and collect data with the Qualtrics survey platform.
We pay workers 2.5~USD per \ratings{} survey and 3~USD per \intrusion{} survey, equivalent to 15~USD/hour.

In order to draw meaningful conclusions from human annotations, we require an adequate number of participants to ensure acceptable statistical power.
However,~\citet{Card2020WithLP} show that many \abr{nlp} experiments, including those relying on human evaluation, are insufficiently powered to detect model differences at reported levels.
Adopting a straightforward generative model of annotations (Appendix~\ref{app:power}), we select enough crowdworkers per task to ensure sufficient statistical power (at least $1-\beta=0.9$) to obtain significance at $\alpha=0.05$, resulting in a minimum of fifteen crowdworkers per topic for both tasks.
On this criterion, both~\citet{chang2009reading} and thus ~\citet{lau2014machine}, with eight annotators, are underpowered.

For each of our two datasets, we generate fifty topics each from the three models in Section~\ref{sec:models:models}.
In the \intrusion{} task, we sample five of the top ten topic words plus one intruder; for the \ratings{} task, we present the top ten words in order (Figure~\ref{fig:tasks}). 
We separate the datasets for each task and randomly sample~40 of the~150 topics. 
In the \ratings{} task, we include an additional sixteen synthetic poor-quality topics to help calibrate scores and filter out low-quality respondents.\footnote{\label{synthetic poor-quality topics}For generating \emph{synthetic poor-quality topics}, we use random high-probability words appearing in topics from other hyperparameter settings, but that have low probability among selected topics. Eight topics each are generated from the vocabularies of \abr{nyt} and \abr{wiki}.}

Phrasing of questions closely follows the wording used by~\citet{chang2009reading}, and crowdworkers received detailed instructions with examples (Appendix~\ref{app:human-eval-instructions}) before responding to items.\footnote{Code to convert topic model output into deployable questionnaires is at \url{github.com/ahoho/topics}.}
As topics can be esoteric (e.g., last columns of Table~\ref{tab:topic_ex}), we ask crowdworkers about their familiarity with the words in each question.
We speculate that this question can help protect against spurious low scores for otherwise coherent topics, as real-world users of topic models are usually familiar with domain-specific terminology (see further discussion in Section~\ref{sec:conclusion}).
\section{Human Judgment Differs From Automated Metrics}\label{sec:analysis}

\begin{figure}
	\centering

	\includegraphics[width=\textwidth]{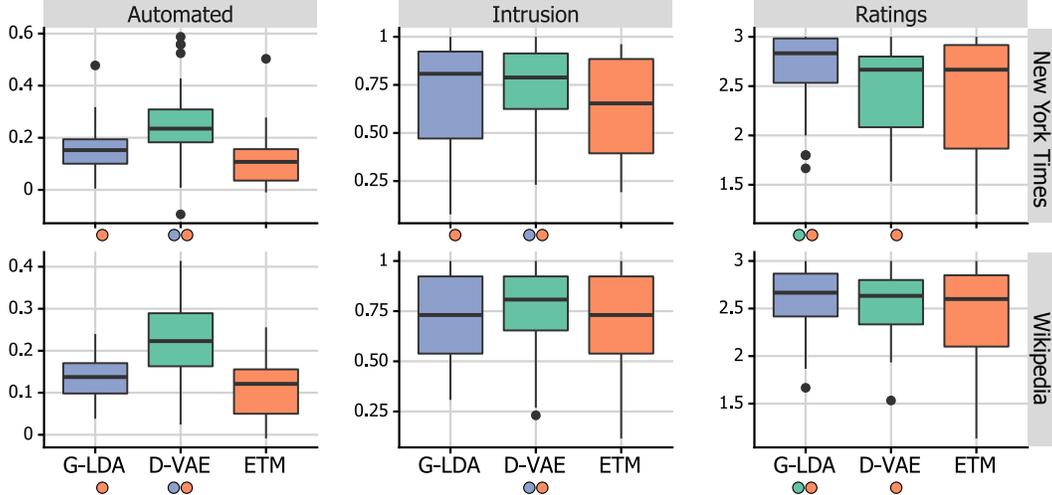}
\caption{While automated evaluations (here, \abr{npmi}) suggest a clear winner between models, human evaluation is more nuanced.
Human judgments exhibit greater variability over a smaller range of values.
Colored circles correspond to pairwise one-tailed significance tests between model scores at $\alpha=0.05$;
for example, the rightmost orange circle at bottom right shows that human intrusion ratings for \abr{d-vae} are significantly higher than \abr{etm} for topics derived from Wikipedia.
}
\label{fig:boxplot}
\end{figure}

 We compare human judgments to automated methods on topics estimated using our three models.

\subsection{Human Assessment}\label{sec:human}

To establish model differences using human ratings, we use pairwise significance tests: a proportion test for the intrusion scores, a \emph{U} test~\citep{mann1947u} for the ratings, and a $t$-test for automated metrics (Fig.~\ref{fig:boxplot}), using one-tailed tests for each pair in both directions.
Although \abr{d-vae} fares better on the intrusion task, evaluation using ratings favors \abr{g-lda}.\footnote{These discrepancies among human tasks support the argument that standard coherence metrics alone may be insufficient for automated model selection~\citep{doogan-buntine-2021-topic}.}

Our human evaluation results are consistent with past iterations of the \ratings{} and \intrusion{} tasks for topic models.  
~\citet{mimno2011optimizing} report an average of 2.36 on the \ratings{} task on a dataset of medical paper abstracts.\footnote{\cite{newman2010automatic} and \cite{lau2014machine} do not report an average.}    
  Our \ratings{} means are 2.5 to 2.8 across all variations (Figure~\ref{fig:boxplot}).
  Our \intrusion{} means range from 0.7 to 0.8, which is comparable to the roughly 0.8 accuracy on the \abr{lda} model evaluated in \citet{chang2009reading}.
 Median time taken on the tasks was 8--9 minutes.

Following~\cite{aletras2013evaluating}, we calculate inter-annotator agreement with the mean Spearman correlation between each respondent's score per topic and the average of other respondent scores, obtaining a value of 0.75 (compare to their value of 0.7 on the \abr{nyt} corpus).
Additionally, we include synthetic poor-quality topics (footnote \ref{synthetic poor-quality topics})---correctly identified by annotators---and we monitor the duration taken for the survey to hedge against insincere submissions.

\subsection{Automated Metrics}\label{sec:auto}

\begin{table}[t]
  \footnotesize
  \begin{center}
  \rowcolors{4}{white}{gray!25}
    \begin{tabular}{ll|rrrr|rrrr}
      \toprule
              &  & \multicolumn{4}{c}{\abr{npmi} (10-token window)} & \multicolumn{4}{c}{$C_v$ (110-token window)} \\
              & Ref. Corpus $\rightarrow$ &  \abr{nyt} &          \abr{wiki} &                  Train &   Val &  \abr{nyt} &          \abr{wiki} &         Train &           Val \\
        & Train Corpus $\downarrow$ &               &                        &                        &       &               &                        &               &               \\
      \midrule
      Intrusion & \abr{nyt} &          0.27 &           \uline{0.43} &                   0.27 &  0.24 &          0.34 &  \textbf{\uline{0.45}} &          0.35 &          0.34 \\
              & \abr{wiki} &  \uline{0.34} &           \uline{0.36} &  \textbf{\uline{0.39}} &  0.17 &  \uline{0.32} &           \uline{0.34} &  \uline{0.34} &          0.20 \\
              & Concatenated &          0.29 &           \uline{0.40} &                   0.32 &  0.17 &          0.32 &  \textbf{\uline{0.40}} &  \uline{0.35} &          0.24 \\
      Rating & \abr{nyt} &          0.37 &  \textbf{\uline{0.48}} &                   0.37 &  0.39 &  \uline{0.41} &           \uline{0.46} &  \uline{0.44} &  \uline{0.45} \\
              & \abr{wiki} &          0.34 &           \uline{0.41} &  \textbf{\uline{0.44}} &  0.28 &          0.32 &           \uline{0.40} &  \uline{0.40} &  \uline{0.34} \\
              & Concatenated &          0.37 &  \textbf{\uline{0.44}} &           \uline{0.41} &  0.35 &          0.38 &           \uline{0.42} &  \uline{0.42} &  \uline{0.42} \\
      \bottomrule
      \end{tabular}
    \end{center}
  \caption{
      Spearman correlation coefficients between mean human scores and automated metrics.
            \uline{Underlined} values have overlapping bootstrapped 95\% confidence intervals with that of the \textbf{largest} value in each row.
            ``Concatenated'' refers to correlations computed on a concatenation of values for the \abr{nyt} and \abr{wiki} items.
            ``Val'' is a small held-out set of 15\% of the training corpus.
            Using the more data-appropriate logistic and ordered probit regressions for \intrusion{} and \ratings{} data leads to different conclusions about relative metric strength (Appendix Table~\ref{tab:auto:logit}).
            CIs are estimated using 1,000 samples.
  }\label{tab:auto}
\end{table}

\abr{npmi} declares \abr{d-vae} the unequivocal victor among the three models (with \abr{g-lda} a clear second), a very different story from the human judgments.
To understand the relationship between automated metrics and human ratings, we estimate the Spearman correlation between the two sets of values for each task and dataset for metric variants (Table~\ref{tab:auto}).
Although previous studies have used mean human ratings over topics, this decision obscures the inherent variance of the human ratings and leads to overconfident estimates.
We therefore construct 95\% confidence intervals by resampling ratings, with replacement, equal to the number of annotators per task (Table ~\ref{tab:auto}). We estimate \abr{npmi} with the standard 10-word window and $C_v$
\citep{roder2015exploring} with the recommended 110-word window.\footnote{We use gensim~\citep{rehurek_lrec} to calculate coherence. We process the reference corpora identically to the training data, retaining only terms that exist in the training vocabulary. Other metrics, like $C_\text{UCI}$~\citep{newman2010automatic} and $C_\text{UMASS}$~\citep{mimno2011optimizing}, show low correlations.}
The Wikipedia corpus appears to be best correlated with human judgments, even for the models trained on the \abr{nyt} corpus---this contradicts~\cite{lau2014machine}, where within-domain data have the highest correlations.

While all correlation coefficients are statistically significant, the strength of the correlation alone does not justify their use in model selection, as is standard in the \abr{ntm} literature (Section ~\ref{sec:meta}).
In particular, the inherent uncertainty of human judgments means that it is difficult to determine when an increase in a model's mean automated coherence implies a significant improvement in the corresponding human scores.\footnote{Better models of human scores could help quantify this relationship (e.g., \textsc{glm}s, see Appendix A.~\ref{tab:auto:logit}).}

As noted above (Figure ~\ref{fig:boxplot}), automated metrics exaggerate model differences compared to human judgments.
To help clarify the utility of automated metrics for model selection, we ask how often an automated metric incorrectly asserts that one model is superior to another.
To do so, we generate a bootstrapped estimate of the false discovery rate of each model.
First, for each dataset, we randomly sample two independent sets of $K=50$ topics (without replacement) from the original pool of 150, along with their corresponding automated and human scores (resampled with replacement, as in Table~\ref{tab:auto}).
Treating the two sampled sets as outputs from two different models, we compute pairwise significance tests between each set for both the $K$ automated metrics and $K\times M$ human scores (using a proportions $z$-test for the intrusion scores and $t$-tests for all other values).
After repeating this process for $N=1000$ iterations, we report the proportion of significant differences detected using the predicted scores despite \emph{equivalent} human scores (after correcting for the probability of type I errors, $\alpha=0.05$).\footnote{Details on testing equivalence are in Section~\ref{app:power:equiv}.}
Even the best-performing automated metrics predict significant differences absent a meaningful human effect roughly one-fifth of the time (Table~\ref{tab:fdr}).

These results suggest that automated metrics alone may be inadequate for model comparison.

\begin{table}[t]
\footnotesize
\begin{center}
  \begin{tabular}{ll|rrr|rrr}
    \toprule
            &  & \multicolumn{3}{c}{\abr{npmi} (10-token window)} & \multicolumn{3}{c}{$C_v$ (110-token window)} \\
            & Ref. Corpus $\rightarrow$ &               \abr{nyt} &              \abr{wiki} &                      Train &               \abr{nyt} &                       \abr{wiki} &                               Train \\
     & Train Corpus $\downarrow$ &                            &                            &                            &                            &                                     &                                     \\
    \midrule
    Intrusion & \abr{nyt} &  46 / \textcolor{gray}{53} &  34 / \textcolor{gray}{48} &  48 / \textcolor{gray}{50} &  35 / \textcolor{gray}{38} &  \textbf{30} / \textcolor{gray}{29} &           34 / \textcolor{gray}{35} \\
            & \abr{wiki} &  44 / \textcolor{gray}{76} &  33 / \textcolor{gray}{78} &  33 / \textcolor{gray}{75} &  45 / \textcolor{gray}{48} &           38 / \textcolor{gray}{49} &  \textbf{37} / \textcolor{gray}{45} \\
            & Concatenated &  42 / \textcolor{gray}{67} &  40 / \textcolor{gray}{66} &  41 / \textcolor{gray}{64} &  36 / \textcolor{gray}{46} &  \textbf{31} / \textcolor{gray}{44} &           30 / \textcolor{gray}{45} \\
    Rating & \abr{nyt} &  45 / \textcolor{gray}{50} &  45 / \textcolor{gray}{51} &  41 / \textcolor{gray}{47} &  27 / \textcolor{gray}{29} &           26 / \textcolor{gray}{26} &  \textbf{21} / \textcolor{gray}{26} \\
            & \abr{wiki} &  40 / \textcolor{gray}{73} &  31 / \textcolor{gray}{73} &  33 / \textcolor{gray}{71} &  38 / \textcolor{gray}{40} &           31 / \textcolor{gray}{40} &  \textbf{28} / \textcolor{gray}{34} \\
            & Concatenated &  39 / \textcolor{gray}{66} &  36 / \textcolor{gray}{66} &  37 / \textcolor{gray}{62} &  31 / \textcolor{gray}{38} &           28 / \textcolor{gray}{38} &  \textbf{19} / \textcolor{gray}{36} \\
    \bottomrule
    \end{tabular}
  \end{center}
\caption{
    False discovery rate (1$-$precision, lower is better) and \textcolor{gray}{false omission rate} of significant model differences when using automated metrics; automated metrics often overstate meaningful model differences.
        \textbf{Bolded} values are those with the lowest geometric mean of FDR and FOR.
        We sample two independent sets of 50 topics along with their human scores and automated metrics; these sets act as the outputs of two ``models''. 
        We then compute significance tests between sets (per Figure~\ref{fig:boxplot}) on both the automated scores and human scores.
        A false positive occurs when one set has significantly larger automated scores despite no meaningful difference in actual human scores.
        Estimates are over 1,000 samples.
}\label{tab:fdr}
\end{table}

\subsection{Explaining the discrepancy}
\begin{figure}
	\centering
	\includegraphics[width=\textwidth]{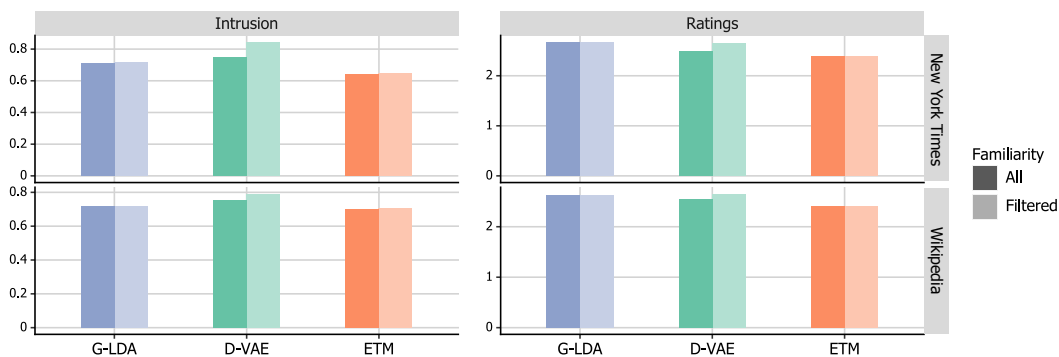}
\caption{Mean human evaluation on the \ratings{} and \intrusion{} tasks, after filtering out respondents who reported a lack of familiarity with the topic words. When filtering, \abr{d-vae} scores improve, highlighting its tendency to produce esoteric topics.}
\label{fig:familiarity}
\end{figure}

\begin{table}[t!]
	\small
	\begin{center}
		\resizebox{\linewidth}{!}{    \begin{tabular}{lllrrr}
      \toprule
      \textbf{Data} & \textbf{Model} &                                                               \textbf{Topic} &  \textbf{\abr{npmi}} &  \textbf{Rat.} &  \textbf{Int.} \\
      \midrule
          \abr{nyt} &    \abr{d-vae} &                                 inc 6mo earns otc rev qtr 9mo nyse outst dec &                 0.56 &           1.60 &           0.77 \\
         \abr{Wiki} &    \abr{d-vae} &   waterline conning turrets boilers amidships aft knots armament guns mounts &                 0.33 &           1.93 &           0.65 \\
         \addlinespace[0.1em]
         \abr{nyt} &    \abr{g-lda} &              bedroom room bath taxes year market listed kitchen broker weeks &                 0.30 &           2.00 &           0.23 \\
          \abr{nyt} &    \abr{d-vae} &       condolences mourns mourn board\_of\_directors heartfelt deepest esteemed &                 0.38 &           2.60 &           0.23 \\
          \midrule[0.06em]
          \abr{nyt} &    \abr{d-vae} &  shareholders earnings federated mci shares takeover new\_york\_stock\_exchange &                 0.18 &           3.00 &           0.81 \\
         \abr{Wiki} &    \abr{d-vae} & continental\_army expedition militia frigate musket frigates muskets skirmish &                 0.11 &           3.00 &           0.69 \\
         \addlinespace[0.1em]
         \abr{nyt} &    \abr{d-vae} &                       medicaid medicare hospitals welfare uninsured patients &                 0.13 &           2.80 &           0.96 \\
          \abr{nyt} &    \abr{g-lda} &  city mayor state new\_york new\_york\_city officials county yesterday governor &                 0.09 &           2.53 &           1.00 \\
      \bottomrule
      \end{tabular}
  }
	\end{center}
	\caption{
    Topics with the largest human--\abr{npmi} discrepancies; top half are topics where \abr{npmi} is high and human preferences are low, bottom half is the reverse. \abr{npmi} favors esoteric and corpus-specific topics. 
    \textbf{\abr{npmi}} is calculated with a 10-token sliding window over the in-domain reference corpus,
		\textbf{Rat.} is the average 3-point rating for a topic, and \textbf{Int.} refers to the percentage of annotators who identify the intruder word.
	}\label{tab:topic-examples}
	
\end{table}

One reason for the discrepancy between human judgments and automated metrics is that metrics favor more esoteric topics.
Specifically, there is a significant negative correlation between a topic's \abr{npmi} or $C_v$ and the share of respondents reporting familiarity with topic words (Pearson's $\rho=-0.29$).
And while \abr{d-vae} achieves the highest automated metric scores of the three models, it produces topics with the fewest familiar words:
respondents report familiarity with terms over 90\% of the time on both tasks for \abr{g-lda} and \abr{etm}, but they do so only 70\% of the time for \abr{d-vae}.
This difference suggests that the topics selected by \abr{d-vae} are narrower in scope than those of the other models.
As shown in Figure~\ref{fig:familiarity}, removing item annotations where respondents indicate unfamiliarity causes both accuracy in the \intrusion{} task and the ratio of ``Very related'' terms in the \ratings{} task for \abr{d-vae} to increase substantially. 
Qualitatively, this result is apparent when examining topics with a high \abr{npmi} but low humans ratings.
In Table~\ref{tab:topic-examples}, the top rows consists of financial terms that frequently appear \emph{together} in \abr{nyt} articles, and the second row contains rare terms about boating---arguably both are reasonable topics for their respective corpora.
We can also see instances where words are qualitatively very related (bottom half of table), but that \abr{npmi} fails to score high---perhaps because these words, while related, may not frequently appear together within a ten-word sliding window (Equation~\ref{eq:npmi}).

Even for familiar words, some topics may be sensible in the context of the specific corpus, despite their component words lacking an immediately obvious semantic relationship.
For example, the topic words in the third and fourth rows appear somewhat unrelated (e.g., ``taxes'' and ``bedroom'' in the third row), but they are in fact characteristic of common document types in the New York \emph{Times}: real estate listings and obituaries.
Topics like these render the \intrusion{} task more difficult: only 23\% of crowdworkers identified the intruder for both topics.

Furthermore, using term familiarity as a proxy for domain expertise does not address the key problems with topic model evaluation:
even after filtering out respondents who are not familiar with topic terms, automated metrics still overstate model differences (Appendix~\ref{app:familiarity}).
The problems with topic model evaluation may therefore extend to our choice of \emph{human} evaluations as well.
\section{So\dots{}is Automated Topic Modeling Evaluation Broken?}
\label{sec:conclusion}

To the extent that our experimentation accurately represents current practice, our results do suggest that topic model evaluation---both automated and human---is overdue for a careful reconsideration.
In this, we agree with \citet{doogan-buntine-2021-topic}, who write that ``coherence measures designed for older models [\dots{}] may be incompatible with newer models'' and instead argue for evaluation paradigms centered on corpus exploration and labeling.
The right starting point for this reassessment is the recognition that both automated and human evaluations are abstractions of a real-world problem.
The familiar use of \mbox{precision-at-10} in information retrieval, for example, corresponds to a user who is only willing to consider the top ten retrieved documents.
In future work, we intend to explore automated metrics that better approximate the preferences of real-world topic model users.

One primary use of topic models is in computer-assisted content analysis.
In that context, rather than taking a methods-driven approach to evaluation, it would make sense to take a needs-driven approach.\footnote{These needs also have a computational component: neural models usually have longer runtimes even when accelerated with GPUs, whereas many practitioners work in local, CPU-only, environments. See Appendix~\ref{app:model-detail} for additional details on runtimes.}
Generic evaluation of topic models using domain-general corpora like NYT needs to be revisited, since there is no such thing as a ``generic'' corpus for content analysis, nor a generic analyst.
\emph{Content analysis} can be formulated in a broad way, as~\cite{kripp2004} has shown, but its actual application is always in a domain, by people familiar with that domain.
This fact stands in tension with the desirable practicalities of general corpora and crowdworker annotation, and the field will need to address this tension.
We have identified ``coherence'' as calling out a latent concept in the mind of a reader.
It follows that we must think about who the relevant human readers are and the conceptual spaces that matter to them.
\section*{Acknowledgements}

This material is based upon work supported by the National Science Foundation under Grants 2031736, 2008761, 1822494, ARLIS, and by an Amazon Research Award.
We thank Sweta Agrawal for her suggestion to conduct a meta-analysis.
We owe much appreciation to Dallas Card for his keen advice on power analyses.
Thanks to Frank Fineis for help on several statistical questions, as well as Shuo Chen for his suggestions regarding the false discovery rate calculations.
Finally, we thank Caitie Doogan for her helpful comments on the clarity of argumentation, as well as our anonymous reviewers.
\bibliography{bib/journal-full,bib/2021_neurips_topics}
\bibliographystyle{style/acl_natbib}

\newpage
\appendix
\section{Appendix}

\subsection{List of Neural Topic Modeling Works used in our Meta-Analysis}\label{app:meta}

\begin{table}
	\centering
	\rotatebox{90}{
		\resizebox{\textheight}{!}{
			\begin{tabular}{lllllllllll}
				\toprule
				Source & Human        & Perplexity & Coherence       & Implementation & Ref. Corpus   & Consistent & Hparam   & >1 run /     & LDA             & Baseline \\
				& Evals?       &            &                 & Specified      & Specified ?   & Preproc?    & search? & err. bars?   & Implementation? & w/in 2 yr?\\
				\midrule
~\cite{Bianchi2020PretrainingIA} & No & No & \abr{npmi}, Embed-sim & None & Internal, External-GoogleNews & Yes & No & Yes & Variational & No\\
~\cite{zhao2021neural} & No & No & \abr{npmi} & Palmetto & No & Unclear & No & Yes & N/A & Yes\\
~\cite{Feng2020ContextRN} & No & Yes & \abr{npmi} & None & No & Yes & No & No & N/A & Yes\\
~\cite{hoyle-etal-2020-improving} & No & No & \abr{npmi} & In paper & External \abr{nyt}, Internal & No & Yes & Yes & N/A & Yes\\
~\cite{Hu2020NeuralTM} & No & No & $C_p$, $C_a$, \abr{npmi} & Palmetto & External \abr{wiki} & No & Likely no & No & Sampling & Yes\\
~\cite{Isonuma2020TreeStructuredNT} & No & Yes & \abr{npmi} & None & No, likely external & Unclear & No & No & Sampling & No\\
~\cite{Joo2020DirichletVA} & No & Yes & \abr{npmi} & None & No, likely internal & No & Likely yes & Yes & N/A & Yes\\
~\cite{Lin2020CopulaGN} & No & Yes & \abr{npmi} & None & No, likely internal & Unclear & Yes & Yes & N/A & Yes\\
~\cite{Ning2020NonparametricTM} & No & Yes & \abr{npmi} & Lau github & No & Yes & Likely no & Yes & Variational & No\\
~\cite{Panwar2020TANNTMTA} & No & No & \abr{npmi} & Lau github & No & Yes & Likely no & No & Sampling & Yes\\
~\cite{Rezaee2020ADV} & No & No & N/A & N/A & N/A & Yes & Likely no & Yes & Variational & No\\
~\cite{Thompson2020TopicMW} & No & No & Coherence, \abr{pmi} & In paper & External \abr{nyt} & No & No & Yes & Sampling & No\\
~\cite{Tian2020LearningVM} & No & Yes & \abr{npmi} & None & No & No & Yes & No & Variational & Yes\\
~\cite{Wang2020NeuralTM} & No & No & $C_p$, $C_a$, \abr{npmi}, UCI & Palmetto & No & No & No & No & Sampling & Yes\\
~\cite{Wu2020NeuralMC} & No & Yes & \abr{npmi} & None & No & No & Yes & No & N/A & Yes\\
~\cite{Wu2020ShortTT} & No & No & $C_v$ & Palmetto & No & Yes & No & No & Unspecified & Yes\\
~\cite{Yang2020GraphAT} & No & Yes & Coherence & In paper & No, likely internal & Yes & No & No & Unspecified & No\\
~\cite{Zhou2020NeuralTM} & No & No & \abr{npmi}, $C_p$ & Palmetto & External \abr{wiki} & No & Likely no & No & Unspecified & Yes\\
~\cite{burkhardtDecouplingSparsitySmoothness2019} & No & Yes & \abr{npmi} & None & No, likely internal & Unclear & Yes & No & Variational & Yes\\
~\cite{diengTopicModelingEmbedding2019} & No & Yes & Coherence & In paper & No, likely internal & Yes & No & No & Unspecified & No\\
~\cite{Gui2019NeuralTM} & No & No & $C_v$ & None & External \abr{wiki} & Yes & Likely no & No & Unspecified & Yes\\
~\cite{Gupta2019DocumentIN} & No & Yes & $C_v$ & Gensim & No, likely internal & Unclear & Likely no & No & N/A & No\\
~\cite{Gupta2019textTOvecDC} & No & Yes & $C_v$ & Gensim & No, likely internal & Unclear & Likely no & No & Sampling & Yes\\
~\cite{Lin2019SparsemaxAR} & No & Yes & PMI & In paper & No, likely external & Unclear & No & No & Variational & Yes\\
~\cite{Liu2019NeuralVC} & No & Yes & \abr{npmi} & Lau github & No, likely internal & Yes & No & No & Variational & Yes\\
~\cite{Nan2019TopicMW} & No & No & \abr{npmi} & None & No & No & No & No & Sampling & Yes\\
~\cite{Wang2019ATMAT} & No & No & $C_p$, $C_a$, UCI, \abr{npmi}, UMASS & Palmetto & No & No & No & No & Unspecified & Yes\\
~\cite{Card2018NeuralMF} & No & Yes & \abr{npmi} & In paper & External-gigaword & Yes & Likely yes & No & Sampling & Yes\\
~\cite{Ding2018CoherenceAwareNT} & No & Yes & \abr{npmi} & Lau github & No, likely external & No & Likely no & No & Sampling & Yes\\
~\cite{He2018InteractionAwareTM} & No & No & Coherence & None & No, likely internal & Yes & No & No & N/A & Yes\\
~\cite{Peng2018NeuralST} & No & Yes & N/A & N/A & N/A & Yes & Likely no & No & Variational & Yes\\
~\cite{Silveira2018TopicMU} & No & Yes & \abr{npmi} & Lau github & Internal & Yes & No & Yes & N/A & Yes\\
~\cite{Zhang2018WHAIWH} & No & Yes & N/A & N/A & N/A & Unclear & Likely no & No & N/A & Yes\\
~\cite{Zhao2018DirichletBN} & No & Yes & \abr{npmi} & Palmetto & External \abr{wiki} & Unclear & No & Yes & N/A & Yes\\
~\cite{Zhu2018GraphBTMGE} & No & No & Coherence & None & No, likely internal & Yes & Likely no & No & Variational & Yes\\
~\cite{Jung2017ContinuousST} & No & Yes & \abr{npmi}, \abr{PMI}, UMASS & None & No & Yes & No & No & Sampling & Yes\\
~\cite{Miao2017DiscoveringDL} & No & Yes & \abr{npmi} & In paper & No & No & Likely no & No & Variational & Yes\\
~\cite{Srivastava2017AutoencodingVI} & No & Yes & \abr{npmi} & None & No & Yes & No & No & Sampling & Yes\\
~\cite{Miao2016NeuralVI} & No & Yes & N/A & N/A & N/A & Yes & Likely no & No & Unspecified & Yes\\
~\cite{Nguyen2015ImprovingTM} & No & No & \abr{npmi} & Lau github & External \abr{wiki} & Yes & No & Yes & Sampling & No\\
				\bottomrule
			\end{tabular}
		}
	}
	\caption{Papers used in meta-analysis, Section~\ref{sec:meta}}\label{tab:meta_full}
\end{table}

In Table~\ref{tab:meta_full}, we report the forty publications used in our meta-analysis (Section~\ref{sec:meta}), which are sourced from a survey of neural topic models~\citep{Zhao2021TopicMM}.

\subsection{Preprocessing Details}\label{app:preprocess}

Our steps are delineated in our implementation,\footnote{\url{github.com/ahoho/topics}} but we list our choices here for easy reference. 
Corpus statistics are in Table~\ref{tab:data}. 
We use the default \texttt{en-core-web-sm} spaCy model~\citep{spacy}, version 3.0.5, throughout.

\begin{table}[h]
	\centering
	\begin{tabular}{l r r}
		\toprule
		& \abr{wiki} & \abr{nyt} \\
		\midrule
				Domain                     & Encyclopedia                            & News                    \\
		\emph{Number of Docs.}     &                                         &                         \\
		\quad Training             & 28.5k                                   & 273.1k                  \\
		\quad Reference            & 4.62M                                   & 1.82M                   \\
		Mean Tokens / Doc.     & 1291                                    & 281                     \\
		Vocab. Size            & 39.7k                                   & 34.6k                   \\
		\bottomrule
	\end{tabular}
	\caption{Corpus statistics. Datasets vary in domain, average document length, and total number of documents. \abr{wiki} is from \cite{merityPointerSentinelMixture2017} and \abr{nyt} is from~\cite{sandhaus2008new}.}
	\label{tab:data}
\end{table}

\begin{itemize}
    \item[] Document processing
    \begin{itemize}
        \item We do not process documents with fewer than 25 whitespace-separated tokens.
        \item Following processing (e.g., stopword removal), we remove documents with fewer than five tokens.
        \item We truncate documents to 5,000 whitespace-separated tokens for \abr{nyt} and to 19,000 for \abr{wiki} (in both cases affecting less than 0.15\% of documents).
    \end{itemize} 
    \item[] Vocabulary creation
    \begin{itemize}
        \item We tokenize using spaCy.
        \item We lowercase terms.
        \item We \emph{do not} lemmatize.
        \item We detect noun entities with spaCy, keeping only the \texttt{ORG}, \texttt{PERSON}, \texttt{FACILITY}, \texttt{GPE}, and \texttt{LOC} types, joining constituent tokens with an underscore (e.g, \mbox{``New York City'' $\rightarrow$ \texttt{new\_york\_city}}).
    \end{itemize}
    \item[] Vocabulary filtering
    \begin{itemize}
		\item The vocabulary is created from the training data. The reference texts used in coherence calculations are processed identically and use the same vocabulary.
        \item We filter out stopwords using the default spaCy English stopword list.\footnote{\url{github.com/explosion/spaCy/blob/v3.0.5/spacy/lang/en/stop_words.py}} Stopwords are retained if they are contained within detected noun entities (e.g., \mbox{$\text{``The United States of America''}\rightarrow\texttt{united\_states\_of\_america}$}).
        \item We filter out tokens with two or fewer characters.
        \item We retain only tokens that are matched by the regular expression \mbox{\texttt{\^{}[\textbackslash{}w-]*[a-zA-Z][\textbackslash{}w-]*\textdollar{}}}
        \item We remove tokens that appear in more than 90\% of documents.
        \item We remove tokens that appear in fewer than $2(0.02|D|)^{1 / \log 10}$ documents, where $|D|$ is the corpus size.\footnote{Standard rules-of-thumb for vocabulary pruning, like removing terms that appear in fewer than 0.5\% of documents~\citep{denny2018text}, ignore the power-law distribution of word frequency~\cite{Zipf49}, and hence do not scale to large corpora. To keep vocabulary sizes roughly consistent across datasets, we set the minimum document-frequency for terms as a (power) function of the total corpus size. This has the intuitive appeal of increasing proportional to the order of magnitude of the number of total documents, starting at a minimum document-frequency of 2 for a 50-document corpus and reaching about 110 for a corpus of 500,000.}
	\end{itemize}
\end{itemize}

\subsection{Training Details}\label{app:model-detail}
Expanding Section~\ref{sec:models:models}, we detail the hyperparameter tuning for each of our three topic models, along with other pertinent details about runtimes and compute resources. 
Scripts used to run the models with all the various hyperparameter configurations are released as part of our code; this section is also included for reference. 

Our general strategy, especially with the neural models, is to select different values around the reported optimal settings in original papers.
For all three models, we try two different values for the number of training iterations (\abr{g-lda}) or epochs (\abr{d-vae}, \abr{etm}).

\paragraph{\abr{g-lda}}
We use gensim~\citep{rehurek_lrec} as a Python wrapper for running Mallet. 
In Table~\ref{tab:glda_hyperparam}, we tune hyperparameters $\alpha$ (topic density parameter) and $\beta$ (word density parameter) which can be thought of as ``smoothing parameters'' that reserve some probability for the topics (words) unassigned to a document (topic) thus far. 
Mallet internally optimizes hyperparameters, and the Optimization Interval controls the frequency of hyperparameter updates, measured in training steps. 

\paragraph{\abr{d-vae}}
Our reimplementation of Dirichlet-\abr{vae}~\citep{burkhardtDecouplingSparsitySmoothness2019} largely uses the same hyperparameters as reported in that work. 
As shown in Table~\ref{tab:dvae_hyperparam}, we vary the prior for the Dirichlet distribution ($\alpha$), the learning rate ($\eta$), the $L_1$-regularization constant for the topic-word distribution~\citep[$\beta_{reg.}$, not in the original model but inspired by ][]{Eisenstein2011SparseAG}, the number of epochs to anneal the use of batch normalization in the decoder~\citep[$\gamma_{BN}$, comes from][]{Card2018NeuralMF}, and the number of epochs to anneal the KL-divergence term in the loss ($\gamma_{KL}$) (it needs to be introduced slowly in the loss function due to the component collapse problem in \abr{VAE}s~\citep{bowman2016generating}). 

\paragraph{\abr{etm}}
Following~\citet{diengTopicModelingEmbedding2019}, we learn skip-gram embeddings on the training corpus using the provided script, which relies on gensim.
As shown in Table~\ref{tab:etm_hyperparam}, we vary the learning rate ($\eta$), the $L_2$ regularization constant for the Adam~\citep{KingmaB14} optimizer ($W_{decay}$), and a boolean indicator of whether to anneal the learning rate ($\gamma_{\eta}$). 
If annealing is allowed, the learning rate gets divided by 4.0 if the loss on the validation set does not improve for more than 10 epochs, per the default settings of the model (preliminary experiments showed that annealing did not attain higher \abr{npmi}).

The runtimes for each of the models on each dataset are in Table~\ref{tab:runtimes}. 
We used AWS ParallelCluster to provide a cloud-computing computing cluster. Neural models ran on NVIDIA T4 GPUs using \texttt{g4dn.xlarge} instances with 16 GiB memory and 4 CPUs.\footnote{\url{https://aws.amazon.com/hpc/parallelcluster/}}
\abr{g-lda} (Mallet) ran on CPU only, with \texttt{m5d.2xlarge} instances (with 32 GiB memory, 8 CPUs).\footnote{See \url{https://aws.amazon.com/ec2/instance-types/} for further details.} 

\begin{table}
    \footnotesize
    \centering
    \begin{subtable}{\textwidth}
		\centering
		\begin{tabular}{c c c c}
			\toprule
			\multicolumn{4}{c}{Model: \textbf{\abr{G-LDA}}} \\
		
		$\alpha$ & $\beta$ & Optim. Interval & $\#\text{Steps}$	                            \\
			\midrule
		$\{0.01, 0.05, 0.1, 0.25^{\dagger}, 1.0^{*}, 5.0\}$ & $\{0.01, 0.05^{*}, 0.1^{\dagger}\}$ & $\{0, 10^{\dagger}, 100, 500^{*}\}$ & $\{1000^{\dagger}, 2000^{*}\}$                 \\
			\bottomrule
		\end{tabular}
		\caption{Hyperparameter ranges for \abr{g-lda}. $\alpha$ is the topic density parameter. $\beta$ is the word density parameter. Optim. Interval sets the number of iterations between Mallet's own internal hyperparameter updates. $\#\text{Steps}$ are training iterations.}
		\label{tab:glda_hyperparam}
    \end{subtable}

    \begin{subtable}{\textwidth}
		\centering
		\resizebox{\textwidth}{!}{
		\begin{tabular}{c c c c c c}
			\toprule
			\multicolumn{6}{c}{Model: \textbf{\abr{d-vae}}} \\
		$\alpha$ & $\eta$ & $\beta_{reg.}$ & $\gamma_{BN}$ & $\gamma_{KL}$ & $\#\text{Steps}$	                        \\
			\midrule
		\shortstack{$\{0.001, 0.01^{*\dagger},  0.1\}$ }& \shortstack{$\{0.001,  0.01^{*\dagger}\}$} & \shortstack{$\{0.0^{*}, 0.01,  0.1^{\dagger}, 1.0\}$} & \shortstack{$\{0, 1^{*},  100, 200^{\dagger}\}$} & \shortstack{$\{100^{*},  200^{\dagger}\}$} & \shortstack{$\{200,  500^{*\dagger}\}$}                 \\
			\bottomrule
		\end{tabular}
		}
		\caption{Hyperparameter ranges for \abr{d-vae}. $\alpha$ is the Dirichlet prior. $\eta$ is the learning rate. $\beta_{reg.}$ is the $L_1$-regularization of the topic-word distribution. $\gamma_{BN}$ and $\gamma_{KL}$ are the number of epochs to anneal the batch normalization constant and KL divergence term in the loss, respectively. $\#\text{Steps}$ are training epochs.}
		\label{tab:dvae_hyperparam}
    \end{subtable}

    \begin{subtable}{\textwidth}
		\centering
		\begin{tabular}{c c c c}
			\toprule
			\multicolumn{4}{c}{Model: \textbf{\abr{etm}}} \\
		
		$\eta$ & $W_{decay}$ & $\gamma_{\eta}$ & $\#\text{Steps}$	                            \\
			\midrule
		$\{0.001^{*}, 0.002, 0.01, 0.02^{*\dagger}\}$ & $\{{1.2e^{-5}}^{*}, {1.2e^{-6}}^{\dagger}, 1.2e^{-7}\}$ & $\{0^{*\dagger}, 1\}$ & $\{500, 1000^{*\dagger}\}$                 \\
			\bottomrule
		\end{tabular}
		\caption{Hyperparameter ranges for \abr{etm}. $\eta$ is the learning rate. $W_{decay}$ is the $L_2$ regularization constant. $\gamma_{\eta}$ is an indicator of whether learning rate is annealed. $\#\text{Steps}$ are training epochs.}
		\label{tab:etm_hyperparam}
    \end{subtable}
\caption{Hyperparameter settings for \abr{g-lda}, \abr{d-vae}, and \abr{etm}. $*$: Best setting for \abr{wiki}, $\dagger$: best setting for \abr{nyt}; based on \abr{npmi} estimated with a 10-token sliding window over the reference corpus.}\label{tab:hyperparams}
\end{table}

\begin{table}
\begin{center}
\begin{tabular}{lrr}
\toprule
  & \textbf{\abr{wiki}} & \textbf{\abr{nyt}}  \\ 
\midrule
\abr{g-lda} & $\sim 2$ minutes & $\sim 9$ minutes  \\ 
\abr{d-vae} & $\sim 45$ minutes & $\sim 330$ minutes   \\ 
\abr{etm} & $\sim 260$ minutes &  $\sim 1300$ minutes  \\ 
\bottomrule
\end{tabular}
\end{center}
\caption{Runtimes for the three topic models on each of the two datasets. \abr{g-lda} requires CPUs only while the neural models use a single GPU. Compute resources detailed at the end of Section~\ref{app:model-detail}.}
\label{tab:runtimes}
\end{table}

\subsection{Instructions for Crowdworkers}\label{app:human-eval-instructions}

Recruiting participants on Prolific.co for a Qualtrics survey produced results with higher inter-worker agreement than Mechanical Turk, based on a pilot test.
Using the Prolific.co platform, we recruited respondents that met the criteria of living in the United States and listing fluency in English.
Each respondent was paid through Prolific upon completion of the survey, at a rate corresponding to \$15 an hour. 
The total amount spent on conducting all the surveys, including our pilot test, was \$2084.91.  
We used automated scripts to generate separate Qualtrics surveys for each task that contained the topics for evaluation, available in our released code.
Each respondent was shown 25\% of the questions in each survey; the question selection and answer display order was chosen randomly via the survey configuration on Qualtrics.
Figures~\ref{fig:tasks:intrusion} and~\ref{fig:tasks:ratings} depict our \intrusion{} and \ratings{} tasks, respectively.
Crowdworkers receive instructions explaining the task (Figure~\ref{fig:instructions}) and the dataset (Figure~\ref{fig:data_descriptions}).

\begin{figure}[t!]
	\centering
		\includegraphics[width=0.8\textwidth]{\figfile{ratings.pdf}}
	\caption{Ratings task presented to crowdworkers.}\label{fig:tasks}\label{fig:tasks:ratings} 
\end{figure}

\begin{figure}[t!]
	\centering
	\begin{subfigure}[t]{0.49\textwidth}
		\centering
		\includegraphics[width=\textwidth]{\figfile{intrusion_instructions.pdf}}
		\caption{}\label{fig:instructions:intrusion}
	\end{subfigure}
	\begin{subfigure}[t]{0.49\textwidth}
		\centering
		\includegraphics[width=\textwidth]{\figfile{ratings_instructions.pdf}}
		\caption{}\label{fig:instructions:ratings}
	\end{subfigure}
	\caption{Instructions for (\subref{fig:instructions:intrusion}) \intrusion{} and (\subref{fig:instructions:ratings}) \ratings{}}\label{fig:instructions}
\end{figure}

\begin{figure}[t!]
	\centering
	\begin{subfigure}[t]{0.49\textwidth}
		\centering
		\includegraphics[width=\textwidth]{\figfile{nytimes_description.pdf}}
		\caption{}\label{fig:data_descriptions:nytimes}
	\end{subfigure}
	\begin{subfigure}[t]{0.49\textwidth}
		\centering
		\includegraphics[width=\textwidth]{\figfile{wiki_description.pdf}}
		\caption{}\label{fig:data_descriptions:wiki}
	\end{subfigure}
	\caption{Descriptions for (\subref{fig:data_descriptions:nytimes}) NYTimes and (\subref{fig:data_descriptions:wiki}) Wikipedia.}
	\label{fig:data_descriptions}
\end{figure}

\subsection{Power Analysis for Human Evaluation Tasks}\label{app:power}

To select the number of crowdworkers, we conduct a power analysis with simulated data~\citep{powerSim} by formulating a generative model of annotations (implementation included in released code).
\cite{Card2020WithLP} find that many \abr{nlp} experiments, including those relying on human evaluation, are insufficiently powered to detect model differences at reported levels.

\paragraph{Word Intrusion.}
Topic $k$ has a true latent binary label $z_k\sim\text{Bern}(0.5)$ (``coherent'' or ``incoherent'') which indexes a parameter $p_{z_k}\in [0,1]$.
Annotator $i$ samples an answer to the intruder task $x_{ik} \sim \text{Bern}(p_{z_k})$. 
We therefore run a simulation of annotator data for two different models: \textsc{model A}, which has a sample of $K=50$ binary topic labels, $\bm{z}^{(A)}$; and \textsc{model B}, with $r$ fewer ``coherent'' topics than \textsc{A}, $\sum_k z^{(B)}_k = \sum_k z^{(A)}_k - r$. 
After collecting pseudo-scores $\bm{x}^{(A)}$ and $\bm{x}^{(B)}$ for $M$ annotators, we run a one-tailed proportion test on the respective sums. 
The power is the proportion of significant tests over the total number of simulations $N$ (i.e., tests there where \textsc{A} is correctly determined to have higher scores than \textsc{B}).
We set $p_0=1/6$ (chance of guessing), $p_1=0.85$~\citep[roughly estimated with data from][]{chang2009reading}.

\paragraph{Ratings.} Rating scores on a 3-point scale are generated analogously, in a generalization of the above binary case.
Assume that topics have true labels $\bm{z}_k\sim\text{Cat}(1/3,1/3,1/3)$.
Annotator scores are noisy, so true labels are corrupted according to probabilities $p_{z_k}\in\Delta^{2}$.
Here, \textsc{model A} has a sample of $K=50$ ratings on a 3-point scale.
\textsc{model B} has $r$ fewer 3-ratings (``very related'') and $r$ greater 1-ratings (``not related'') than \textsc{A} (the 2-ratings stay constant).
After simulating scores for $M$ annotators for both ``models,'' we run a one-tailed \emph{U}-test~\citep{mann1947u}.
Again, the power is the share of significant tests over all simulations $N$.
Probabilities are $p_1=[3/4,1/4,0];p_2=[1/4,2/4,1/4];p_3=[0,1/4,3/4]$, designed to roughly approximate empirical data---if we sample scores according to them and compute inter-``annotator'' agreement, the one-versus-rest Spearman correlation is $\rho\approx0.7$, or the same as the most-correlated dataset (\abr{nyt}) in \cite{aletras2013evaluating} (our final data has $\rho=0.75$).

For both settings, we set $r=4$, the critical value $\alpha=0.05$, and the desired power $1-\beta=0.9$. This analysis suggests fifteen annotators per topic for the ratings task and twenty-five for intrusion. 

\subsubsection{Power analysis for equivalence}\label{app:power:equiv}

To estimate the false discovery (omission) rates in Table~\ref{tab:fdr}, we need to determine when differences between human (automated) scores are not meaningful.
Since human effects in the opposite direction of automated metrics also imply a false discovery, we conduct a test of non-inferiority; this is the same as using a large negative lower bound in the two-one-sided tests procedure for equivalence~\citep{schuirmann1987comparison,wellek2010testing}.

To determine the non-inferiority threshold---the bound $\epsilon$ below which we consider two sets of scores to be equivalent---we also conduct a power analysis, per the previous section.
In this case, the simulation assumes \emph{no} difference between the ``true'' labels of the model outputs, $\bm{z}^{(A)}=\bm{z}^{(B)}$.
We estimate one-sided tests for each sample of human scores, with the null~$H_0: \mu_1^{(B)} - \mu^{(A)} > \epsilon$ for some bound $\epsilon$.
We minimize $\epsilon$ while maintaining $\beta>0.9$. This process produces $\epsilon=0.05$ for the \intrusion{} task and $\epsilon=0.11$ for the \ratings{} task (roughly equivalent to a difference of 2.5 ``incoherent'' topics for both tasks, respectively).

For the automated scores, we generate two sets of scores $x_k \sim \mathcal{N}(0, \sigma^2);\ \sigma^2 \sim \text{Gamma}(\alpha, \beta)$ for $k=1\ldots K$ at each iteration, then conduct a t-test between each set. $\alpha$ and $\beta$ are selected such that the Gamma distribution approximately matches the empirical distribution of automated score variances. This leads to $\epsilon=0.05$ for \abr{npmi} scores and $\epsilon=0.06$ for the $C_v$ scores.

\subsection{Regression Results}\label{app:regression}
\begin{table}[t]
    \footnotesize
    \begin{center}
    \rowcolors{4}{white}{gray!25}
    \begin{tabular}{ll|rrrr|rrrr}
        \toprule
                &  & \multicolumn{4}{c}{\abr{npmi} (10-token window)} & \multicolumn{4}{c}{$C_v$ (110-token window)} \\
                & Ref. Corpus $\rightarrow$ &  \abr{nyt} &          \abr{wiki} &                  Train &   Val &  \abr{nyt} & \abr{wiki} &                  Train &           Val \\
         & Train Corpus $\downarrow$ &               &                        &                        &       &               &               &                        &               \\
        \midrule
        Intrusion & \abr{nyt} &          2.42 &  \textbf{\uline{4.16}} &                   2.11 &  1.97 &          2.50 &  \uline{3.27} &                   2.55 &          2.40 \\
                & \abr{wiki} &  \uline{4.11} &           \uline{5.08} &  \textbf{\uline{5.45}} &  0.87 &          2.23 &          2.79 &                   2.74 &          0.70 \\
                & Concatenated &          2.82 &  \textbf{\uline{4.56}} &                   3.18 &  0.78 &          2.30 &          3.05 &                   2.64 &          0.87 \\
        Rating & \abr{nyt} &          1.92 &           \uline{2.08} &                   1.77 &  1.85 &  \uline{2.55} &  \uline{2.51} &  \textbf{\uline{2.68}} &  \uline{2.59} \\
                & \abr{wiki} &          2.97 &           \uline{4.10} &  \textbf{\uline{4.29}} &  1.45 &          2.01 &          2.82 &                   2.86 &          0.80 \\
                & Concatenated &  \uline{2.20} &  \textbf{\uline{2.75}} &           \uline{2.52} &  1.17 &  \uline{2.27} &  \uline{2.60} &           \uline{2.74} &          1.07 \\
        \bottomrule
    \end{tabular}
	\end{center}
    \caption{
        Logistic (intrusion) and ordinal probit (ratings) regression coefficients of automated metrics on human annotations.
                \uline{Underlined} values have overlapping 95\% confidence intervals with that of the \textbf{largest} value in each row.
    }\label{tab:auto:logit}
    \end{table}

Prior work~\cite[e.g.,][]{roder2015exploring} relates averaged human ratings to automated metrics using either Pearson or Spearman correlations.
As an alternative that takes into account both the variation in human judgments as well as their numerical type, we estimate logistic and ordered probit regressions on the ratings and intrusion annotations, respectively.
In Table~\ref{tab:auto:logit}, we report the estimated coefficients for each metric, finding that---on the whole---using the \abr{wiki} reference performs best, although the large estimated confidence intervals mitigate the strength of this conclusion.

\subsection{Filtering on Term Familiarity}\label{app:familiarity}
\begin{table}
    \footnotesize
    \centering
    \begin{subtable}{\textwidth}
        \centering
        \rowcolors{4}{white}{gray!25}
		\begin{tabular}{ll|rrrr|rrrr}
			\toprule
					&  & \multicolumn{4}{c}{\abr{npmi} (10-token window)} & \multicolumn{4}{c}{$C_v$ (110-token window)} \\
					& Ref. Corpus $\rightarrow$ &  \abr{nyt} &          \abr{wiki} &         Train &   Val &  \abr{nyt} &          \abr{wiki} &         Train &   Val \\
			 & Train Corpus $\downarrow$ &               &                        &               &       &               &                        &               &       \\
			\midrule
			Intrusion & \abr{nyt} &          0.34 &           \uline{0.51} &          0.32 &  0.25 &          0.44 &  \textbf{\uline{0.55}} &          0.42 &  0.38 \\
					& \abr{wiki} &  \uline{0.39} &           \uline{0.39} &  \uline{0.40} &  0.14 &  \uline{0.38} &  \textbf{\uline{0.40}} &  \uline{0.39} &  0.13 \\
					& Concatenated &          0.36 &           \uline{0.45} &          0.36 &  0.18 &          0.41 &  \textbf{\uline{0.48}} &          0.41 &  0.26 \\
			Rating & \abr{nyt} &          0.45 &  \textbf{\uline{0.59}} &          0.44 &  0.43 &          0.51 &           \uline{0.58} &  \uline{0.53} &  0.52 \\
					& \abr{wiki} &  \uline{0.45} &  \textbf{\uline{0.51}} &  \uline{0.51} &  0.21 &  \uline{0.44} &           \uline{0.51} &  \uline{0.51} &  0.23 \\
					& Concatenated &          0.47 &  \textbf{\uline{0.54}} &          0.47 &  0.35 &  \uline{0.49} &           \uline{0.53} &  \uline{0.51} &  0.42 \\
			\bottomrule
			\end{tabular}
    \caption{
        Spearman correlation coefficients between mean human scores and automated metrics, compare to~Table~\ref{tab:auto}.
    }\label{tab:auto_familiar}
    \end{subtable}

    \begin{subtable}{\textwidth}
        \centering
		\begin{tabular}{ll|rrr|rrr}
			\toprule
					&  & \multicolumn{3}{c}{\abr{npmi} (10-token window)} & \multicolumn{3}{c}{$C_v$ (110-token window)} \\
					& Ref. Corpus $\rightarrow$ &               \abr{nyt} &              \abr{wiki} &                      Train &                        \abr{nyt} &                       \abr{wiki} &                               Train \\
				& Train Corpus $\downarrow$ &                            &                            &                            &                                     &                                     &                                     \\
			\midrule
			Intrusion & \abr{nyt} &  53 / \textcolor{gray}{55} &  46 / \textcolor{gray}{50} &  56 / \textcolor{gray}{52} &           41 / \textcolor{gray}{34} &  \textbf{28} / \textcolor{gray}{27} &           42 / \textcolor{gray}{33} \\
					& \abr{wiki} &  38 / \textcolor{gray}{76} &  36 / \textcolor{gray}{77} &  32 / \textcolor{gray}{76} &  \textbf{29} / \textcolor{gray}{37} &           31 / \textcolor{gray}{43} &           33 / \textcolor{gray}{39} \\
					& Concatenated &  54 / \textcolor{gray}{70} &  41 / \textcolor{gray}{73} &  51 / \textcolor{gray}{70} &           41 / \textcolor{gray}{44} &  \textbf{29} / \textcolor{gray}{42} &           41 / \textcolor{gray}{44} \\
			Rating & \abr{nyt} &  45 / \textcolor{gray}{49} &  39 / \textcolor{gray}{53} &  45 / \textcolor{gray}{47} &           18 / \textcolor{gray}{27} &  \textbf{16} / \textcolor{gray}{24} &           17 / \textcolor{gray}{25} \\
					& \abr{wiki} &  37 / \textcolor{gray}{73} &  25 / \textcolor{gray}{74} &  30 / \textcolor{gray}{70} &           28 / \textcolor{gray}{31} &           19 / \textcolor{gray}{33} &  \textbf{18} / \textcolor{gray}{27} \\
					& Concatenated &  45 / \textcolor{gray}{64} &  38 / \textcolor{gray}{68} &  42 / \textcolor{gray}{64} &           26 / \textcolor{gray}{36} &  \textbf{21} / \textcolor{gray}{36} &           27 / \textcolor{gray}{33} \\
			\bottomrule
			\end{tabular}
    \caption{
        False discovery rate (1$-$precision, lower is better) and \textcolor{gray}{false omission rate} of significant model differences when using automated metrics, compare to~Table~\ref{tab:fdr}.
    }\label{tab:fdr_familiar}
    \end{subtable}

    \begin{subtable}{\textwidth}
        \centering
        \rowcolors{4}{white}{gray!25}
        \begin{tabular}{ll|rrrr|rrrr}
        \toprule
                &  & \multicolumn{4}{c}{\abr{npmi} (10-token window)} & \multicolumn{4}{c}{$C_v$ (110-token window)} \\
                & Ref. Corpus $\rightarrow$ &  \abr{nyt} &          \abr{wiki} &         Train &   Val & \abr{nyt} & \abr{wiki} & Train &   Val \\
        & Train Corpus $\downarrow$ &               &                        &               &       &              &               &       &       \\
        \midrule
        Intrusion & \abr{nyt} &          3.71 &  \textbf{\uline{7.14}} &          3.04 &  2.54 &         3.34 &          4.54 &  3.23 &  2.94 \\
                & \abr{wiki} &  \uline{5.87} &  \textbf{\uline{6.46}} &  \uline{6.19} &  0.85 &         3.23 &          3.59 &  3.39 &  0.42 \\
                & Concatenated &          4.24 &  \textbf{\uline{6.81}} &          4.17 &  0.94 &         3.18 &          4.06 &  3.30 &  0.91 \\
        Rating & \abr{nyt} &          4.40 &  \textbf{\uline{5.87}} &          3.85 &  3.93 &         3.97 &          4.44 &  4.03 &  3.89 \\
                & \abr{wiki} &  \uline{4.84} &  \textbf{\uline{5.95}} &  \uline{5.65} &  1.33 &         2.96 &          3.73 &  3.69 &  0.62 \\
                & Concatenated &          4.49 &  \textbf{\uline{5.80}} &          4.56 &  1.78 &         3.45 &          3.91 &  3.81 &  1.32 \\
        \bottomrule
        \end{tabular}
    \caption{
        Logistic (intrusion) and ordinal probit (ratings) regression coefficients of automated metrics on human annotations, compare to Table~\ref{tab:auto:logit}.
    }\label{tab:auto:logit_familiar}
    \end{subtable}
\caption{Tables~\ref{tab:auto},~\ref{tab:fdr}, and~\ref{tab:auto:logit} after removing respondents who report a lack of familiarity with topic words.}\label{tab:filter_on_familiar}
\end{table}

Several topics, particularly those produced by \abr{d-vae}, contain terms that are not well-known to annotators (\ref{sec:human}).
When a respondent is unfamiliar with a topic's words, their ratings for that topic may not accurately reflect its true coherence.
For example, a mycologist may find the words in the fifth column of Table~\ref{tab:topic_ex} highly related, whereas someone unfamiliar with fungi-related jargon may rate it poorly---indeed, the mean rating for this topic is 2.1 for those unfamiliar with terms and 2.6 for those who are familiar. 

Since automated metrics do not take into account a term's familiarity to humans, we posit that automated metrics should be more predictive of human judgments among respondents who are familiar with topic terms.
To test this hypothesis, we re-evaluate the relationships between automated metrics and human judgments \emph{after removing} respondents who state they are not familiar with a topic's terms (Table~\ref{tab:filter_on_familiar}).
On the whole, results are much clearer than above; \abr{npmi} estimated using \abr{wiki} reference counts is strongly correlated across tasks and datasets.
The false discovery rate is lower overall, although automated metrics still misdiagnose significant results at a rate of one in six in even the best case.

These findings provide further evidence---per our discussion in Section~\ref{sec:conclusion}---that future human evaluations of topic models ought to take into account domain expertise and information need.

\subsection{Five-point Ratings Scale}
\begin{table}[t]
    \footnotesize
    \begin{center}
    \rowcolors{4}{white}{gray!25}
	\begin{tabular}{ll|rrrr|rrrr}
		\toprule
				  &  & \multicolumn{4}{c}{\abr{npmi} (10-token window)} & \multicolumn{4}{c}{$C_v$ (110-token window)} \\
				  & Ref. Corpus $\rightarrow$ & \abr{nyt} &          \abr{wiki} & Train &           Val & \abr{nyt} & \abr{wiki} &         Train &                    Val \\
		 & Train Corpus $\downarrow$ &              &                        &       &               &              &               &               &                        \\
		\midrule
		Rating (5-pt.) & \abr{nyt} &         0.27 &  \textbf{\uline{0.37}} &  0.28 &  \uline{0.33} &         0.29 &  \uline{0.35} &  \uline{0.33} &           \uline{0.35} \\
				  & \abr{wiki} &         0.15 &                   0.21 &  0.29 &  \uline{0.43} &         0.10 &          0.16 &          0.17 &  \textbf{\uline{0.50}} \\
				  & Concatenated &         0.21 &                   0.30 &  0.28 &          0.32 &         0.20 &          0.26 &          0.26 &  \textbf{\uline{0.39}} \\
		\bottomrule
		\end{tabular}
	\end{center}
    \caption{
        Spearman correlation coefficients between mean human scores for a \textbf{five-point} ratings scale (rather than three), compare to~Table~\ref{tab:auto}.
                \uline{Underlined} values have overlapping 95\% confidence intervals with that of the \textbf{largest} value in each row.
    }\label{tab:auto:five-point}
    \end{table}

Although most prior work uses three-point scales for the \ratings task (Fig. \ref{fig:tasks:ratings}), for comparison we also ask annotators to label the topic topic words with a five-point scale ranging from 1 (``not at all related'') to 5 (``very related'', no labels are given for points 2-4). Broadly, we find that values for correlations are reduced relative to the three-point scale (Table~\ref{tab:auto:five-point}). We believe examining this discrepancy is an interesting direction for future work that re-visits human evaluation of topic models.

\subsection{Potential Negative Impact}\label{app:impact}
Our work focuses its investigation on data from the English language alone. In this way, it further entrenches English-language primacy in \abr{nlp}, and more crucially, findings may not translate directly to other languages. We caution the reader against applying claims made in this work to topic modeling on corpora of other languages. It is even possible that one of the tasks designed to elicit human judgment (e.g., \intrusion{}) may not be amenable for use with other languages. 

Concerning topic models more broadly, we note that others question the scholarly value of ``distant reading'' and the digital humanities in general~\citep{marche2012humanities,allington2016humanities}. Do topic models encourage a passive, disengaged relationship to texts---fomenting conclusions about broad, generic trends rather than idiosyncratic specifics, leading us to miss the trees for the forest? As noted by~\cite{schmidt2012words}, ``topics neither can nor should be studied independently of a deep engagement in the actual word counts that build them.'' In this light, topic models can be viewed as an extension of the insidious neoliberal trend toward mass data harvesting that blurs differences between individuals and cultures. Researchers should take care to avoid such elisions when drawing conclusions from model outputs.

\end{document}